\documentclass[10pt,twocolumn,letterpaper]{article}

\usepackage{wacv}
\usepackage{times}
\usepackage{epsfig}
\usepackage{graphicx}
\usepackage{amsmath}
\usepackage{amssymb}

\wacvfinalcopy 

\def\assignedStartPage{1} 

\usepackage{booktabs}
\usepackage{authblk}
\usepackage{pifont}
\makeatletter
\@namedef{ver@everyshi.sty}{}
\makeatother
\usepackage{tikz}
\usetikzlibrary{arrows,positioning}
\usepackage[final]{microtype}
\usepackage{hyphenat}

\usepackage{adjustbox}
\usepackage{todonotes}

\ifwacvfinal
\usepackage[breaklinks=true,bookmarks=false]{hyperref}
\else
\usepackage[pagebackref=true,breaklinks=true,colorlinks,bookmarks=false]{hyperref}
\fi

\ifwacvfinal
\else
\fi

\hypersetup{
    pdfstartview={FitH},
    pdftitle={Find it if You Can: End-to-End Adversarial Erasing for Weakly-Supervised Semantic Segmentation},
    pdfauthor={Erik Stammes, Tom F.H. Runia, Michael Hofmann, Mohsen Ghafoorian},
    pdfsubject={Find it if You Can: End-to-End Adversarial Erasing for Weakly-Supervised Semantic Segmentatio},
    pdfcreator={Erik Stammes},
    pdfproducer={Erik Stammes},
    pdfkeywords={Deep Learning} {Semantic Segmentation} {Weak Supervision} {Adversarial Erasing},
}

\begin{document}

\title{\emph{Find it if You Can:} End-to-End Adversarial Erasing \\ for Weakly-Supervised Semantic Segmentation}

\author[1,2]{Erik Stammes}
\author[1]{Tom F.H. Runia}
\author[2]{Michael Hofmann}
\author[2]{Mohsen Ghafoorian}
\affil[1]{University of Amsterdam, Amsterdam, the Netherlands \authorcr Email: {\tt erikstammes@me.com, tomrunia@gmail.com}\vspace{1.5ex}}
\affil[2]{TomTom, Amsterdam, the Netherlands \authorcr Email: {\tt mohsen.ghafoorian@tomtom.com} \vspace{-2ex}} 

\maketitle
\begin{abstract}
Semantic segmentation is a task that traditionally requires a large dataset of pixel-level ground truth labels, which is time-consuming and expensive to obtain.
Recent advancements in the weakly-supervised setting show that reasonable performance can be obtained by using only image-level labels. 
Classification is often used as a proxy task to train a deep neural network from which attention maps are extracted.
However, the classification task needs only the minimum evidence to make predictions, hence it focuses on the most discriminative object regions. 
To overcome this problem, we propose a novel formulation of adversarial erasing of the attention maps.
In contrast to previous adversarial erasing methods, we optimize two networks with opposing loss functions, which eliminates the requirement of certain suboptimal strategies;
for instance, having multiple training steps that complicate the training process or a weight sharing policy between networks operating on different distributions that might be suboptimal for performance.
The proposed solution does not require saliency masks, instead it uses a regularization loss to prevent the attention maps from spreading to less discriminative object regions.
Our experiments on the Pascal VOC dataset demonstrate that our adversarial approach increases segmentation performance by 2.1 mIoU compared to our baseline and by 1.0 mIoU compared to previous adversarial erasing approaches.
\end{abstract}


\begin{figure}[t]
\centering
\begin{adjustbox}{max width=\textwidth}
\begin{tabular}{c c}
\includegraphics[trim={3cm 0 0 0},clip,width=3.5cm]{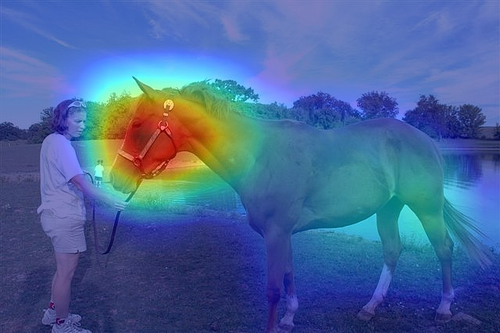} & \includegraphics[trim={3cm 0 0 0},clip,width=3.5cm]{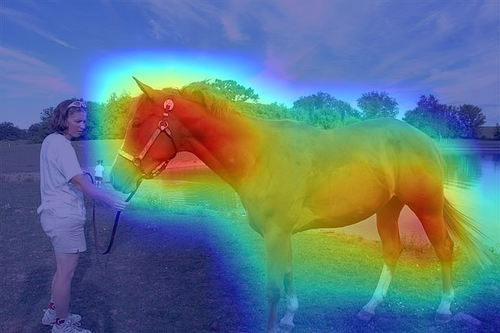}\\
\includegraphics[trim={0 0 0 0},clip,width=3.5cm]{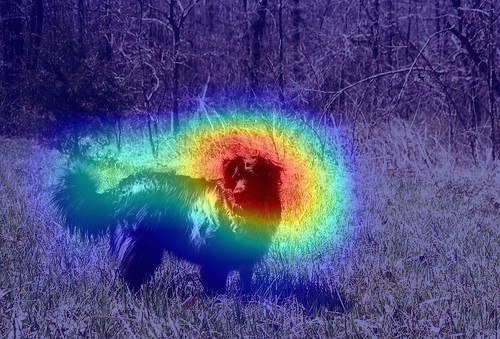} &
\includegraphics[trim={0 0 0 0},clip,width=3.5cm]{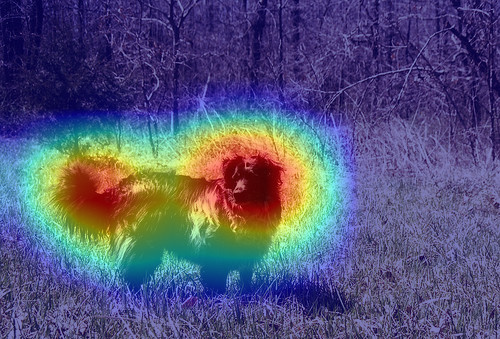}

\end{tabular}
\end{adjustbox}
\caption{Two examples of attention maps obtained from a classification network (left) and end-to-end adversarial erasing (right). Classification networks need only the minimum evidence to classify the objects that are present, hence they focus on the most discriminative objects regions. We call this the \textit{discriminative localization problem}. Our proposed end-to-end adversarial erasing scheme resolves this problem by spreading the attention to less discriminative object regions.}
\label{fig:discriminative}
\end{figure}

\tikzstyle{block} = [draw, rectangle, 
    minimum height=2em, minimum width=4em, node distance=2cm]
\tikzstyle{sum} = [draw, circle, minimum size=1cm, node distance=2cm]
\tikzstyle{dia} = [diamond, draw, text badly centered, inner sep=2t]
\tikzstyle{square} = [draw, rectangle, 
    minimum height=2em, minimum width=2em, node distance=2cm]
\tikzstyle{input} = [coordinate]
\tikzstyle{output} = [coordinate]
\tikzstyle{pinstyle} = [pin edge={to-,thin,black}]
\begin{figure*}[!t]
\centering
\begin{adjustbox}{max width=1\textwidth}
\begin{tikzpicture}[auto,  node distance=1.4cm,>=latex']
    \node[name=input2, inner sep =0, outer sep=0]{\includegraphics[height=1cm]{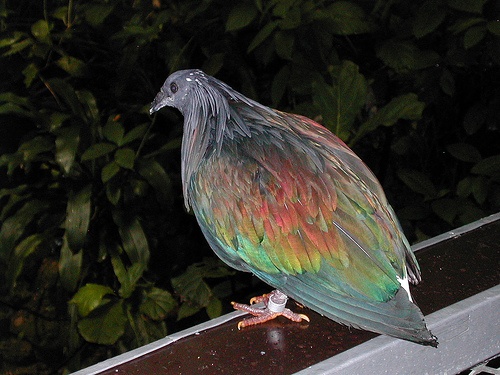}};
    \node[name=input2_erased, below=0.34cm of input2, inner sep =0, outer sep=0]{\includegraphics[height=1cm]{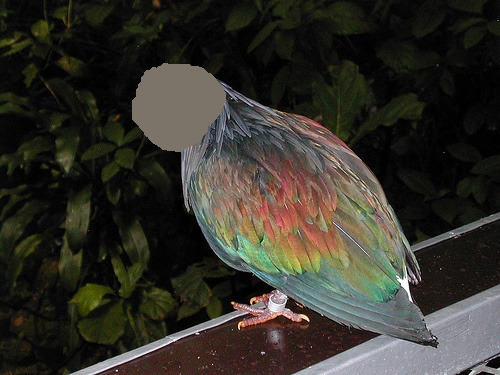}};
    \node[draw, rectangle, node distance=0.4cm, minimum height=1.0cm, minimum width = 1.0cm,  right= 1.1cm of input2, name=cnn3]{CNN};
    \node[circle, draw=black!80, below right=0cm and 0.3cm of input2, name=minus2]{$-$};
    \node[draw, rectangle, node distance=0.4cm, minimum height=1.0cm, minimum width = 1.0cm,right=1.1cm of input2_erased, name=cnn4]{CNN};
    \node[right=0.25cm of cnn3, inner sep =0, outer sep=0, name=attention2]{\includegraphics[height=1cm]{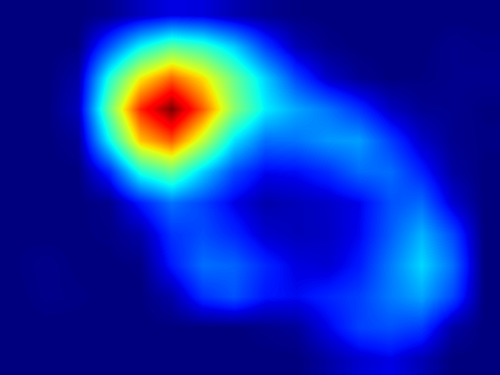}};
    \node[right=0.25cm of cnn4, inner sep =0, outer sep=0, name=attention3]{\includegraphics[height=1cm]{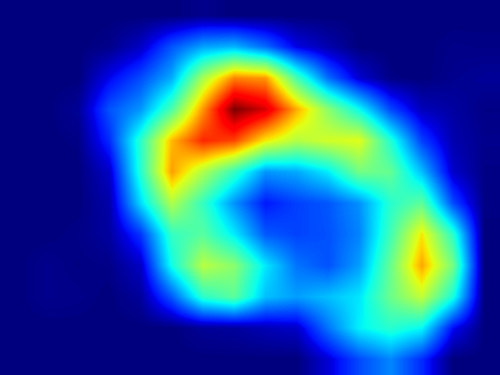}};
    \node[circle, draw=black!80, right=0cm and 3cm of minus2, name=fusion]{$+$};
    \node[right=0.25cm of fusion, inner sep =0, outer sep=0, name=fused_attention]{\includegraphics[height=1cm]{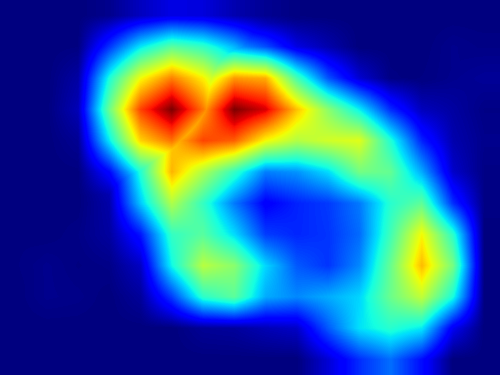}};
    \node[below=0cm of cnn4, name=dots]{{\huge \vdots}};
    
    \draw[-latex] (input2) to (cnn3);
    \draw[-latex] (input2) to (minus2);
    \draw[-latex] (minus2) to (input2_erased);
    \draw[-latex] (input2_erased) to (cnn4);
    \draw[-latex] (cnn3) to (attention2);
    \draw[-latex] (cnn4) to (attention3);
    \draw[-latex] (attention2.south west) to [out=210, in=10] (minus2);
    \draw[-latex] (attention2.east) to (fusion);
    \draw[-latex] (attention3.east) to (fusion);
    \draw[-latex] (fusion) to (fused_attention);

    \node[right=0.75cm of fused_attention, inner sep =0, outer sep=0, name=input]{\includegraphics[height=1cm]{imgs/comparison-figure/img.jpg}};
    \node[draw, rectangle, node distance=0.4cm, minimum height=1.0cm, minimum width = 1.0cm, right= 0.25cm of input, name=cnn1]{CNN};
    \node[right=0.25cm of cnn1, inner sep =0, outer sep=0, name=attention1] {\includegraphics[height=1cm]{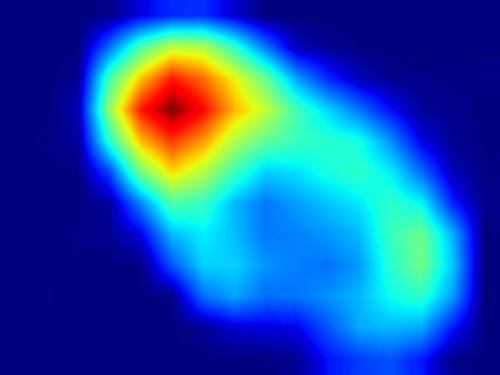}};
    \node[circle, draw=black!80, right= 0.25cm of attention1, name=minus1]{$-$};
    \node[right=0.25cm of minus1, inner sep =0, outer sep=0, name=erased]{\includegraphics[height=1cm]{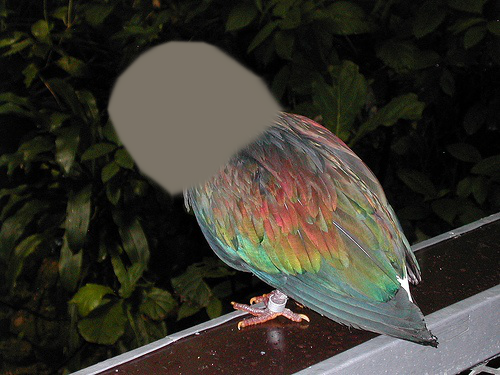}};
    \node[draw, rectangle, node distance=0.4cm, minimum height=1.0cm, minimum width = 1.0cm, align=center, right=0.25cm of erased, name=cnn2]{Adv.\\ CNN};
    
    \draw[-latex] (input) to (cnn1);
    \draw[-latex] (cnn1) to (attention1);
    \draw[-latex] (attention1) to (minus1);
    \draw[-latex] (input.south east) to [out=310, in=240] (minus1.south west);
    \draw[-latex] (minus1) to (erased);
    \draw[-latex] (erased) to (cnn2);


\end{tikzpicture}
\end{adjustbox}
\caption{High-level comparison of the iterative adversarial erasing methods~\cite{wei2017object,zhang2018adversarial} (left) to our novel proposed approach (right) trained with a single iteration of adversarially trained models.} 
\label{fig:comparisonl}
\end{figure*}

\section{Introduction}
Semantic segmentation is among the most fundamental tasks in computer vision, with applications ranging from autonomous vehicles~\cite{siam2017deep} to medical diagnosis~\cite{litjens2017survey}.
There has been remarkable progress in quality of semantic segmentation models in the era of deep learning~\cite{badrinarayanan2017segnet,chen2018encoder,zhao2017pyramid}, in part due to the availability of large-scale datasets with pixel-level ground truth labels~\cite{everingham2010pascal,Geiger2013IJRR,lin2014microsoft}. 
However, labeling these datasets with pixel-level annotations is a laborious process. 
Weakly-supervised methods achieve reasonable performance with much coarser labels such as bounding boxes~\cite{dai2015boxsup,xu2015learning}, scribbles~\cite{lin2016scribblesup,wu2018scribble}, points~\cite{bearman2016s} or even image-level labels~\cite{ahn2018learning,wei2017object,zhang2019reliability}.
In this work we focus on leveraging image-level labels, which are the weakest form of supervision. 
It is common in methods that use only image-level labels to train a classification network and extract class activation maps (CAMs) as initial object locations~\cite{ahn2018learning,jiang2019integral,wei2018revisiting}. 
However, learning semantic segmentation with only image-level labels is an ill-posed problem, since the labels indicate only the existence of a class instead of its location and shape. More specifically, the attended visual evidence generally corresponds to the most discriminative object regions and therefore fails to capture the complete object~\cite{zhou2016learning,li2018tell,zhang2018adversarial}. 
We call this the \textit{discriminative localization problem}, which is illustrated in the left images of Figure~\ref{fig:discriminative}.
This problem is especially prevalent in non-rigid object classes such as birds, cats, horses and sheep where the texture of the fur or skin is much less discriminative than other body parts such as heads or feet.

Previous methods~\cite{zhang2018adversarial,wang2018weakly,li2018tell,hou2018self,wei2017object} propose to alleviate this problem by introducing \textit{adversarial erasing}, which sets a threshold on the attention map to generate a mask which can be used to remove the most discriminative object regions from the image. 
The resulting image is then fed into a second classification network to find less discriminative regions that belong to the same object.
Some of the existing methods perform the erasing in multiple steps, either implemented as a multi-stage training approach~\cite{wei2017object} or trained in an integrated fashion with multiple erasing networks trained jointly~\cite{zhang2018adversarial}. 
This will result in either a complicated multi-stage training strategy or a more extensive memory footprint that might hinder leveraging state-of-the-art network architectures.
Figure~\ref{fig:comparisonl} illustrates a high-level schematic comparison between the existing iterative erasing methods~\cite{zhang2018adversarial,wei2017object} and our proposed end-to-end approach. 
Other methods~\cite{li2018tell} aim at avoiding this shortcoming by training a single erasing step while sharing the weights among the models operating on the input and erased input. 
The weight sharing, however, might result in suboptimal performance given the different distributions of data they are operating on. 

Our proposed method follows the adversarial erasing methodology to recover the less discriminative object regions, but in contrast, we propose to train two separate networks, a localizer network and an adversarial network, in a truly adversarial manner.
By involving the two networks in an adversarial game, we encourage the localizer networks to leave no visual clues for the adversarial networks to discover the existence of the corresponding class.
Moreover, we regularize the localizer network to prefer solutions with smaller attention maps, to avoid low-specificity localization solutions that cover more than necessary in favor of winning the adversarial game. 
As a result, compared to previous methods, this setup eliminates the need for multiple consecutive localizer models during training and inference~\cite{wei2017object,zhang2018adversarial} and weight sharing between models that operate on different data distributions~\cite{li2018tell}. 
Furthermore, our proposed framework does not rely on extra supervision such as additional data or saliency estimation~\cite{wei2017object,li2018tell,wang2018weakly,hou2018self}.
To demonstrate the effectiveness of our method we not only show improved results with the proposed adversarial training scheme as a stand-alone model, but also integrate our end-to-end adversarial erasing in Pixel-Level Semantic Affinity (PSA)~\cite{ahn2018learning} and achieve better segmentation performance.

The main contributions of this paper are as follows: (1)  we propose a novel end-to-end adversarial erasing method which helps capturing less discriminative object regions, (2) we show how this approach can be integrated into existing weakly-supervised semantic segmentation methods, and (3) we demonstrate its effectiveness on the Pascal VOC 2012 benchmark, outperforming the baselines. The implementation is included as supplementary material and will be made publicly available upon the acceptance of the paper.

\section{Related work}
\paragraph{Visual Attention.}
Since the early days of the breakthrough of deep neural networks, considerable attention was given to shed light into these ``black boxes'' to better understand the decision making process.
For instance, in visual tasks such as image classification it is often useful to highlight the image regions responsible for the network's decision. 
Earlier work achieved this by visualizing partial derivatives of predicted class scores w.r.t. the input image~\cite{simonyan2013deep} or by making modifications to raw gradients~\cite{zeiler2014visualizing}. 
CAMs~\cite{zhou2016learning} can highlight relevant regions by adapting a global average pooling layer and a fully connected layer for classification. 
Selvaraju \emph{et al.}~\cite{selvaraju2017grad} extend this approach to Grad-CAM which utilizes gradients to make it possible to get visual explanations for tasks such as image captioning and visual question answering without any network architecture changes.

\paragraph{Adversarial Erasing.}
Visual attention techniques are often used for downstream tasks such as object detection and semantic segmentation when there is only a weak supervision signal~\cite{ahn2018learning,zhang2018adversarial}. 
Often, classification is used as a proxy task to generate the attention maps. 
Since classification needs only the minimum evidence to make a prediction, only the most discriminative regions of an image are used in the decision making process.
In downstream tasks this leaves unsatisfactory results, as the goal is to capture the entire object in the image.
To mitigate this issue, adversarial erasing was first introduced by~\cite{wei2017object}.
In adversarial erasing, the most discriminative object region is found using attention maps and then erased from the images.
The erased images are then sent into another classification network to find less discriminative object regions belonging to the same entity. 
Finally, the attention maps are combined to create a segmentation mask.
This approach has been improved by~\cite{zhang2018adversarial}, which integrated the erasing step into training by erasing from the feature map instead of the image. 
However, both of these approaches still require multiple training and/or inference steps and the fusion of attention maps into segmentation masks. 
Li \emph{et al.}~\cite{li2018tell} resolve this by sharing weights between two classifiers and applying a soft thresholding technique, which allows the attention maps of the initial classifier to grow to less discriminative object regions~\cite{li2018tell}. 
In adversarial erasing approaches, the attention often starts to spread to the background regions that are highly correlated with the corresponding objects.
This can be fixed by using extra supervision in the form of saliency masks~\cite{hou2018self}.
In contrast to the previous approaches, we utilize adversarial erasing without the need of multiple localizer models during training/inference, weight sharing or saliency masks.
We achieve this by training two models using distinct optimizers with adversarial objectives.
Our approach is simple and can be easily plugged into existing weakly-supervised methodologies.

\paragraph{Weakly-Supervised Semantic Segmentation.}
In weakly-supervised semantic segmentation (WSSS), the supervision signal is reduced from pixel-level labels to bounding boxes~\cite{dai2015boxsup,xu2015learning}, scribbles~\cite{lin2016scribblesup,wu2018scribble}, points~\cite{bearman2016s} or even image-level labels~\cite{ahn2018learning,wei2017object,zhang2019reliability}.
In this work we focus on image-level labels, as it is the hardest task and reduces the labeling efforts the most.
A number of methods utilize adversarial erasing to produce semantic segmentation masks~\cite{wei2017object,li2018tell,hou2018self}.
Furthermore, there are methods that randomly hide parts of the feature map~\cite{lee2019ficklenet,choe2019attention} and methods that utilize cross-image features~\cite{wei2016learning,fan2018associating,fan2018cian,wang2020deep}.
In many weakly-supervised methods class agnostic saliency methods are used as cues of object and background~\cite{wei2018revisiting,jiang2019integral,wei2016learning,fan2018associating,fan2018cian,lee2019ficklenet,wei2017object,li2018tell,hou2018self}.
Common to many WSSS methods, the output segmentation masks are used as proxy labels to train a fully supervised semantic segmentation model~\cite{wei2017object,li2018tell,hou2018self,lee2019ficklenet,zhang2019reliability}.
Our framework does not require saliency masks and is agnostic to the choice of training a fully supervised semantic segmentation model.

\paragraph{Adversarial Training.}
The idea of adversarial training has gained significant attention in recent years after the introduction of Generative Adversarial Nets (GANs)~\cite{goodfellow2014generative}, consisting of two competing networks, the generator and the discriminator. 
The task of the discriminator is to predict whether a given input image is coming from the real data distribution or the distribution of fake images generated by the generator, while the task of the generator is to fool the discriminator by matching the distribution of real data. 
This approach to image synthesis has been proven to be powerful and has resulted in generating convincing looking images~\cite{karras2019style,karnewar2020msg}. 
The idea of adversarial training has since been extended to different tasks, such as image-to-image translation~\cite{isola2017image}, reconstructing 3D objects from images~\cite{wu2016learning}, image super-resolution~\cite{wang2018esrgan} and semantic segmentation~\cite{luc2016semantic,samson2019bet}. 
Similar to our approach, adversarial training has been used to find complete segmentation masks from weak supervision~\cite{shetty2018adversarial}, but unlike our method this setup imposes shape priors on the generator and is used for the task of automatic object removal. 
The term adversarial training has also been loosely used in the WSSS field, where it denotes erasing part of an image and training an auxiliary model on this new image, despite not having any adversarial objective formulation and/or independent and competing models with different parameterization.
Our approach to adversarial training is closer to the original adversarial training formulation, as we use two distinct models with opposing objectives. 

\section{Method}
In this section the proposed method, \textbf{e}nd-to-end \textbf{ad}versarial \textbf{er}asing (EADER), is described. First, we present our novel adversarial training formulation for weakly-supervised semantic segmentation.
Then we illustrate the effectiveness of our method, by integrating end-to-end adversarial erasing into an existing weakly supervised semantic segmentation framework.

\subsection{End-to-End Adversarial Erasing}
\label{sec:adv}
\begin{figure*}[!t]
\centering
\begin{tikzpicture}[auto,  node distance=1.4cm,>=latex']
    \node[inner sep=0pt, name=x, label={[yshift=-0cm]$x$}](x) {\includegraphics[height=1.2cm ]{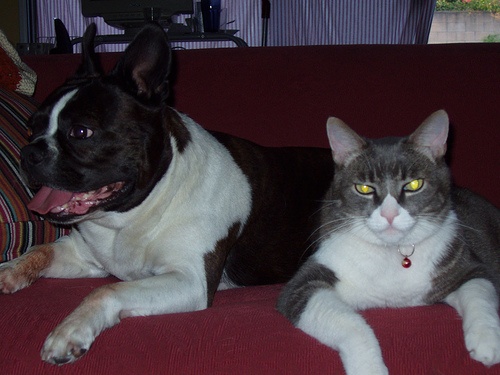}};
    \node[text width = 2cm, align=center, below = 0.15cm of x, name=y](y) {$y$: \footnotesize{\{cat, dog\}}};
    \node[draw, rectangle, minimum height=1.2cm, minimum width = 0.05cm, right = 0.2cm of x, name=enc1]{};
    \node[draw, rectangle, node distance=0.4cm, minimum height=1.0cm, minimum width = 0.01cm, right of=enc1, name=enc2]{};
    \node[draw, rectangle, node distance=0.4cm, minimum height=0.75cm, minimum width = 0.01cm, right of=enc2, name=enc3, label={[label distance=0.3cm]localizer ($G_\varphi$)}]{};
    \node[draw, rectangle, node distance=0.4cm, minimum height=0.5cm, minimum width = 0.01cm, right of=enc3, name=enc4]{};
    \node[name=cam_cat, above right= -0.5cm and 0.9cm of enc4,  inner sep =0, outer sep=0, label={[yshift=-0cm]attention map (A$_c$)}](cam_cat) {\includegraphics[height=1.2cm ]{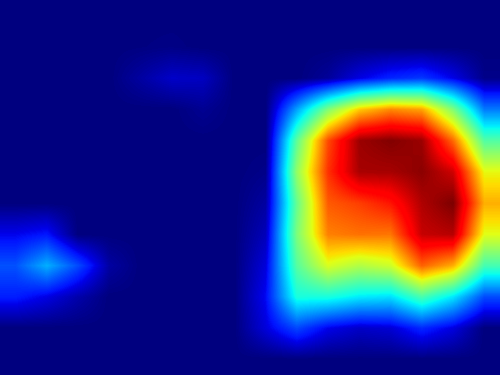}};
    \node[ name=cam_dog, right= 0.4 cm of enc4,  inner sep =0, outer sep=0](cam_dog) {\includegraphics[height=1.2cm]{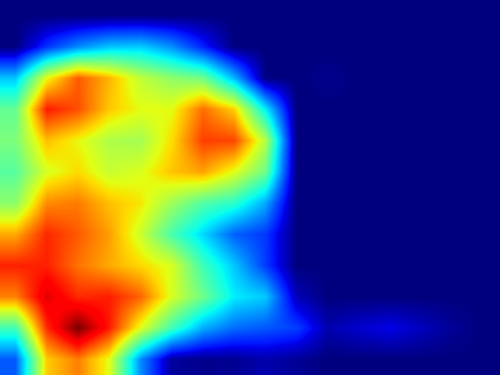}};
    \node (loss_loc) [text width=2cm, align=left, below right= 0.4cm and -0.4cm of enc4, name=loss_loc] {$\mathcal{L}_\text{loc}$ {\scriptsize w.r.t.} $\varphi$};
    \node[circle, draw=black!80, right =0.8cm of cam_dog, name=thresh](thresh) {\includegraphics[height=0.3cm]{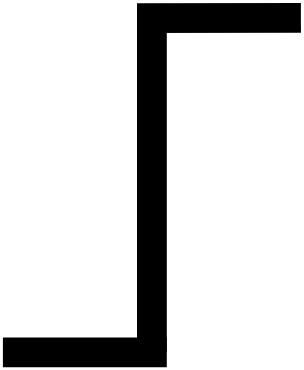}};
    \node[name=mask_cat, above right= -0.5cm and 0.9cm of thresh,  inner sep =0, outer sep=0, label={[yshift=-0cm]mask (M$_c$)}](mask_cat) {\includegraphics[height=1.2cm]{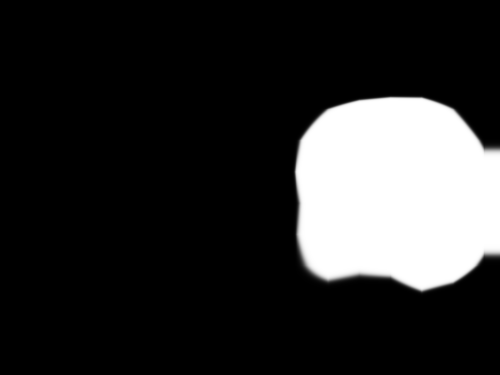}};
    \node[name=mask_dog, right= 0.4cm of thresh,  inner sep =0, outer sep=0](mask_dog) {\includegraphics[height=1.2cm]{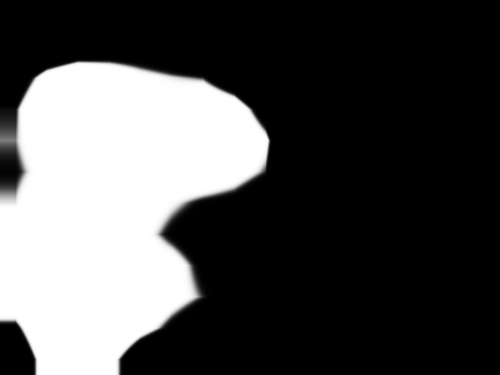}};
    \node[text width=2cm, align=left, below right=-0.2cm and 0cm of cam_dog, name=loss_reg](loss_reg) {$\mathcal{L}_\text{reg}$ {\scriptsize w.r.t.} $\varphi$};
    \node[circle, draw=black!80, below =0.2cm of loss_loc, name=minus](minus) {$-$};
    \node[name=erased_cat, below right= 1.3cm and 0.2cm of cam_cat,  inner sep =0, outer sep=0, label={[yshift=-0.1cm]$\tilde{x}$}] (erased_cat) {\includegraphics[height=1.2cm]{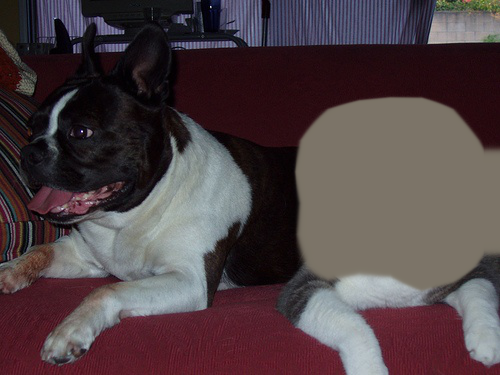}};
    \node[name=erased_dog, below right= 1.3cm and 0.2cm of cam_dog,  inner sep =0, outer sep=0] (erased_dog) {\includegraphics[height=1.2cm]{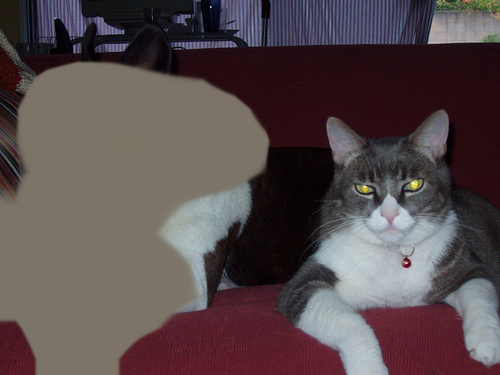}};
    \node[draw, rectangle, minimum height=1.2cm, minimum width = 0.05cm, right = 0.2cm of erased_cat, name=adv1]{};
    \node[draw, rectangle, node distance=0.4cm, minimum height=1.0cm, minimum width = 0.01cm, right of=adv1, name=adv2]{};
    \node[draw, rectangle, node distance=0.4cm, minimum height=0.75cm, minimum width = 0.01cm, right of=adv2, name=adv3, label={[label distance=0.2cm]adversarial ($F_\theta$)}]{};
    \node[draw, rectangle, node distance=0.4cm, minimum height=0.5cm, minimum width = 0.01cm, right of=adv3, name=adv4]{};
    \node[text width=2cm, align=left, below right=-0.2cm and 0.4cm of adv4, name=loss_adv](loss_adv) {$\mathcal{L}_\text{adv}$ {\scriptsize w.r.t.} $\theta$};
    \node[text width=2cm, align=left, above right=-0.2cm and 0.4cm of adv4, name=loss_am](loss_am) {$\mathcal{L}_\text{am}$ {\scriptsize w.r.t.} $\varphi$};
    \node[circle, draw=black!80, below left =1cm and -0.1cm of x, name=leg_minus](leg_minus) {$-$};
    \node[circle, draw=black!80, below =0.2cm of leg_minus, name=leg_thresh](leg_thresh) {\includegraphics[height=0.3cm]{imgs/threshold.png}};
    \node[text width=2.5cm, align=left, right= 0.1 cm of leg_minus, name=text_minus](text_minus)  {= erasing};
    \node[text width=3.5cm, align=left, right= 0.1 cm of leg_thresh, name=text_thresh](text_thresh)  {= soft thresholding};

    \draw[-latex] (x) to node[above,rotate=0] {} (enc1);
    \draw[-latex] (enc1) to node[above,rotate=0] {} (enc2);
    \draw[-latex] (enc2) to node[above,rotate=0] {} (enc3);
    \draw[-latex] (enc3) to node[above,rotate=0] {} (enc4);
    \draw[-latex] (enc4) to node[above,rotate=0] {} (cam_dog);
    \draw[-latex] (enc4) to node[above,rotate=0] {} (loss_loc);
    \draw[-latex] (mask_dog.south) to [out=200, in=0] (minus);
    \draw[-latex] (x.south east) to [out=330, in=180] (minus);
    \draw[-latex] (minus) to (erased_dog);
    \draw[-latex] (cam_dog) to (thresh);
    \draw[-latex] (thresh) to (mask_dog);
    \draw[-latex] (erased_cat) to (adv1);
    \draw[-latex] (adv1) to (adv2);
    \draw[-latex] (adv2) to (adv3);
    \draw[-latex] (adv3) to (adv4);
    \draw[-latex] (adv4) to (loss_am);
    \draw[-latex] (adv4) to (loss_adv);
    \draw[-latex] (cam_dog.east) to (loss_reg);
 
\end{tikzpicture}
\caption{An overview of our end-to-end adversarial erasing framework. The images $x$ are forwarded through the localizer $G_\varphi$ to extract per-class ($c$) attention maps $A_c$. Using a soft-thresholding operation they are converted to masks $M_c$, which are used to create images where the most discriminative object parts have been erased ($\tilde{x}$). These are forwarded through the adversarial $F_\theta$, which is optimized using a classification loss $\mathcal{L}_\text{adv}$. The localizer is optimized using a classification loss $\mathcal{L}_\text{loc}$ and an adversarial loss term $\mathcal{L}_\text{am}$. This forces the localizer to spread its attention to less discriminative object parts, while the $\mathcal{L}_\text{reg}$ loss encourages the model to bound the activation to the minimum necessary area.} 
\label{fig:model}
\end{figure*}

Our proposed method consists of two image classifiers: an image classifier and an adversarial model. 
Both can be instantiated by any appropriate convolutional neural network.
The first image classifier network is used to localize the target object using attention maps, hence we call this network the \textit{localizer network}. 
The attention maps are then converted to masks by a soft, differentiable thresholding operation. 
Next, the masks are used to create a new image where the the most discriminative object regions are erased.
These images are then forwarded through the second network, which we call the \textit{adversarial network}. 
Its goal is to classify the images correctly, even when the target classes are erased. 
The localizer and adversarial networks' image classifiers are optimized using binary cross entropy loss, but in alternating fashion using distinct optimizers.
To force the localizer network to not only classify the image correctly, but also to spread its attention to less discriminative object regions, we add an adversarial loss term to the localizer.
This term captures the ability of the adversarial network to still classify the erased object.
A trivial solution for the localizer would then be to hide the entire image from the adversarial, hence we regularize the localizer with an additional regularization loss term. 
This limits the attention of the localizer and thus forces it to only erase the regions that belong to the target class. An overview of the end-to-end adversarial framework is shown in Figure~\ref{fig:model}.

\tikzstyle{block} = [draw, rectangle, 
    minimum height=2em, minimum width=4em, node distance=2cm]
\tikzstyle{sum} = [draw, circle, minimum size=1cm, node distance=2cm]
\tikzstyle{dia} = [diamond, draw, text badly centered, inner sep=2t]
\tikzstyle{square} = [draw, rectangle, 
    minimum height=2em, minimum width=2em, node distance=2cm]
\tikzstyle{input} = [coordinate]
\tikzstyle{output} = [coordinate]
\tikzstyle{pinstyle} = [pin edge={to-,thin,black}]

 Consider a dataset $\mathcal{D} = \left\{ x_i, y_i \right\}^N_{i=1}$, where
 $N$ is the number of images, $x_i$ the input image and $y_i$ a multi-hot vector of length $C$, with $C$ being the number of classes, and in which $y_{i,c}=1$, if class $c$ is present in $x_i$ and $y_{i,c}=0$, otherwise. Note that being in a multi-label setup, multiple classes can be present in an input image and hence $\sum_{c}{y_{i,c}} \geq 1$.
 
\paragraph{Localizer network.}
The localizer can be instantiated by any convolutional neural network from which (Grad-)CAMs can be extracted. For simplicity we assume the usage of CAMs but it is straightforward to extend this approach to more advanced attention extraction methods such as Grad-CAM. The localizer $G$ with trainable parameters $\varphi$ is trained as multi-label classifier using binary cross entropy loss on each label class: 

\begin{align}
\begin{split}
    \mathcal{L}_\text{loc}\left(G_{\varphi}(x_i), y_{i}\right) = -\frac{1}{C}\sum_{c}y_{i,c}\  \text{ln}\left(G_{\varphi}(x_i)\right)\\
    +(1-y_{i,c})\ \text{ln}\left(1-G_{\varphi}(x_i)\right)
\end{split}
\end{align}

\paragraph{Attention maps.}
Given a trained localizer network $G_\varphi$, the attention map $A_c$ for class $c$ can be obtained using its feature map of the final convolutional layer $g_\varphi^{\text{final}}$ and the classification weights $w_c$ as follows:
\begin{align}
    A_c(x_i) = \text{ReLU}\left(w_c^T g_\varphi^{\text{final}}(x_i)\right).
\end{align}
$A_c$ is then normalized so that the maximum activation equals 1.
\paragraph{Soft masks.}
Only the attention maps for ground truth classes are kept, which are then resized to the input image dimensions and a soft thresholding operation is applied to generate class specific masks $M_c$
\begin{align}
\label{eq:mask}
    M_c(x_i) = \sigma\left(\omega\left(\text{A}_c(x_i) - \psi\right)\right),
\end{align} 
where $\sigma$ is the sigmoid non-linearity, $\psi$ is the threshold value and $\omega$ is a scaling parameter that ensures that values above the threshold are (close to) 1 and values below are (close to) 0. 
In contrast to a regular thresholding operation, this soft threshold is differentiable which allows the gradients from any further computations to backpropagate to the localizer. 

\paragraph{Erasing.}
The input images for the adversarial network, where the attention maps have been erased, are computed as follows:
\begin{align}
    \tilde{x}_{i,c} = x_i \odot (1 - M_c(x_i))
\end{align}
Note here that only the attention map of one particular class is erased, which is why multiple images are created in cases where there is more than one target. 

\paragraph{Adversarial network.}
The adversarial network $F$ with trainable parameters $\theta$ is then trained as multi-label classifier using the same binary cross entropy loss function:
\begin{align}
\label{eq:adv_cls}
\begin{split}
    \mathcal{L}_\text{adv}\left(F_\theta(\tilde{x}_i), y_i\right) = -\frac{1}{C} \sum_{c} y_{i,c}\  \text{ln}\left(F_\theta(\tilde{x}_{i, c})\right)\\
    +(1-y_{i,c})\ \text{ln}\left(1-F_\theta(\tilde{x}_{i,c})\right)
\end{split}
\end{align}
Hence, the goal of this network is to classify the same targets as before, despite the erased evidence.

\paragraph{Attention mining loss.}
To encourage the model to erase the object evidence thoroughly, we engage the localizer network in an adversarial game with the adversarial model. We follow~\cite{li2018tell} by utilizing attention mining loss, which is the mean of the logits of the classes that have been erased:
\begin{align}
    \mathcal{L}_\text{am}(\tilde{x}_i, y_i) = \frac{1}{C} \sum_{c \in y_i} F_\theta(\tilde{x}_{i,c})
\end{align}

\paragraph{Regularization loss.}
Finally, to regularize the localizer, we impose an additional loss term:
\begin{align}
    \mathcal{L}_\text{reg}(x_i, y_i) = \frac{1}{W \times H \times C} \sum_{c \in y_i} \sum_{j, k} A_c(x_i)_{j,k},
\end{align}
where $W$, and $H$ represent the width and height of the activations.
Incorporating this regularization loss in the optimization process encourages the localizer to find a \emph{minimum} attention map that covers the target class and hence prevents the localizer from the trivial solution where it erases the entire image to globally minimize the attention mining loss.

\paragraph{Total loss.}
The total loss function for the localizer then becomes:
\begin{align}
\label{eq:totalloss}
    \mathcal{L}_\text{total} = \mathcal{L}_\text{loc} + \alpha \mathcal{L}_\text{am} + \beta \mathcal{L}_\text{reg},
\end{align}
where $\alpha$ and $\beta$ are hyper-parameters to tune the importance of the adversarial and regularization losses respectively.
While the localizer is trained to minimize its adversarial loss term, the adversarial model tries to maximize it, by minimizing its loss in Equation~\ref{eq:adv_cls}.

\paragraph{Segmentation maps.}
After training the model with the described loss terms, we convert the attention maps to segmentation maps. 
We first upsample and stack all the attention maps into the image resolution with $C+1$ channels. Since we do not train the classification models for the background class, we set the first channel to a threshold value of $\rho$. 
To obtain the segmentation masks we take the argmax over the class dimension.

\subsection{Integrability of End-to-End Adversarial Erasing}
\label{sec:int-psa}
The proposed method is simple and integrable, and we showcase this by integrating the proposed end-to-end adversarial erasing scheme into an existing WSSS method. 
We integrate it into Pixel-level Semantic Affinity (PSA)~\cite{ahn2018learning}, a multi-stage method which suffers from the discriminative localization problem in its first stage. 
In this stage a classification network is trained from which CAMs are extracted. 
This stage does not utilize specific methods to improve the segmentation masks, but training and test-time data augmentations increase performance in this regard. 
The latter two stages train \textit{AffinityNet}, which generates pseudo segmentation masks, and a fully-supervised segmentation model, which uses the pseudo masks as training data.

The CAM-generation stage of PSA is suitable for adversarial training as it suffers from the discriminative localization problem and because the classification network is suitable as a localizer, i.e. the CAMs are generated from the final convolutional layer without the need of any post-processing or other gradient-breaking computations.
As before, we apply a soft threshold on the attention maps to create masks, which are then used to erase the most discriminative object regions from the input images. 
The resulting images are forwarded through the adversarial network and attention mining loss is applied as adversarial loss on the localizer network.

Note that a baseline method need not be multi-stage to make it suitable for integrability of EADER. 
As long as the attention map can be obtained without breaking the gradients, EADER can be integrated into the method to find less discriminative object regions to improve the attention maps.

\section{Experiments}
\subsection{Experimental Setup}
\paragraph{Dataset.} We evaluate the performance of the proposed method on the Pascal VOC 2012 segmentation dataset~\cite{everingham2010pascal}, the most widely used benchmark on weakly supervised semantic segmentation.
The dataset consists of 20 object classes and one background class and contains 1464, 1449 and 1456 images in the train, validation and test sets respectively. Following the previous works in the WSSS literature, we augment the dataset with annotations from Hariharan \emph{et al}.~\cite{BharathICCV2011}, resulting in a total of 10582 training images. We report the mean intersection-over-union (mIoU) for the validation and test sets. The test set results are obtained using the official Pascal VOC evaluation server.

In contrast to the previous adversarial erasing methods we do not employ any post-processing and keep the tricks to a minimum to keep our method simple. More specifically, we leave out tricks such as test-time augmentations, post-processing and saliency cues to the method we integrate with. 

\paragraph{Network architecture details.} 
We test the adversarial training approach with a ResNet-101~\cite{he2016deep} localizer network, while the adversarial model is a ResNet-18 network.
We utilize ImageNet pre-trained weights for both networks. 
When integrating with PSA, to ensure fair comparisons, we do not change any of the existing networks, which means the localizer is a WideResNet~\cite{wu2019wider} with 38 convolutional layers, while the new adversarial model is a ResNet-18. 
In the final stage we train a fully supervised semantic segmentation network on proxy labels. 
We utilize DeepLabV3+, which is a modern segmentation model, with ResNet-101 and Xception-65 backbones and the default training strategy from~\cite{chen2018encoder}.

\paragraph{Training specifications.}
We train the localizer with a batch size of 16, while the batch size is dynamic for the adversarial model as it depends on the number of objects in each image. For example, when each image in the batch of 16 has two object classes, both objects are erased from each image separately and the batch size for the adversarial network will be 32.
We randomly resize and crop the input images into 448 $\times$ 448 for both the localizer and the adversarial model.
Both networks are optimized for 10 epochs with stochastic gradient descent with a learning rate of 0.01. 
We alternately train the localizer and adversarial per 200 training steps.
Throughout the experiments, unless specified otherwise, we have used an $\alpha$ value of 0.05 and $\beta$ is set to $10^{-5}$ (Equation~\ref{eq:totalloss}). Further hyperparameter values are $\omega = 100$, $\psi = 0.5$ (both Equation~\ref{eq:mask}) and  $\rho = 0.3$.
We follow the training settings of~\cite{ahn2018learning} when training PSA and use an initial learning rate of 0.01 for the adversarial network. 
To show that our method is agnostic to attention map generation method, we utilize Grad-CAM~\cite{selvaraju2017grad} in our experiments and CAM~\cite{zhou2016learning} when integrating into PSA.
This also ensures a fair comparison to PSA, which utilizes CAM.

\subsection{Ablation study}

\begin{table}[t]
\centering
\setlength{\tabcolsep}{0.5em}
{\renewcommand{\arraystretch}{1.2}
\begin{tabular}{@{}rccc@{}}
\toprule
$\alpha$ & mIoU         & Precision    & Recall       \\ \midrule
0     & 41.37 {\scriptsize $\pm$ 0.26} & 58.26 {\scriptsize $\pm$ 0.50} & 58.18 {\scriptsize $\pm$ 0.66} \\
0.01  & 42.51 {\scriptsize $\pm$ 0.41} & 57.79 {\scriptsize $\pm$ 0.72} & 60.88 {\scriptsize $\pm$ 0.52} \\
0.05  & \textbf{43.89} {\scriptsize $\pm$ 0.40} & 54.78 {\scriptsize $\pm$ 1.03} & 68.13 {\scriptsize $\pm$ 1.51} \\
0.1   & 42.88 {\scriptsize $\pm$ 0.99} & 52.68 {\scriptsize $\pm$ 2.30} & 69.31 {\scriptsize $\pm$ 1.84} \\ \bottomrule
\end{tabular}
}
\caption{Performance of the model with different $\alpha$ values on the Pascal VOC 2012 validation set. We report both the mean and standard deviation over 6 runs.}
\label{tab:alpha-ablation}
\end{table}

\begingroup
\setlength{\tabcolsep}{1pt} 
\renewcommand{\arraystretch}{0.5} 
\begin{figure}[t]
\centering
\begin{adjustbox}{max width=\hsize}
\begin{tabular}{c c c c}
\includegraphics[width=2cm]{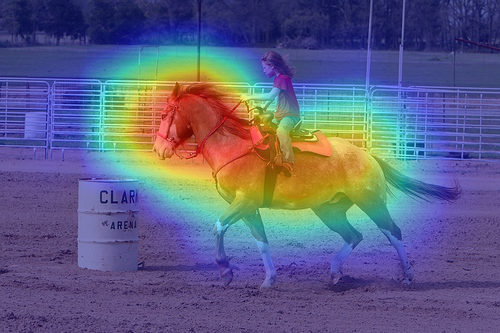}&\includegraphics[width=2cm]{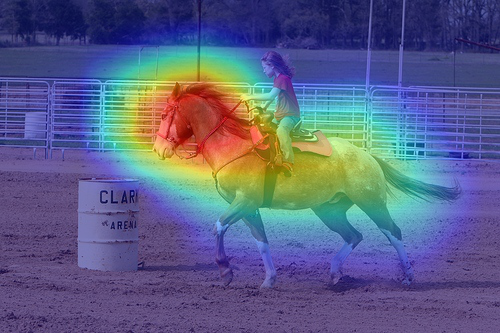}&\includegraphics[width=2cm]{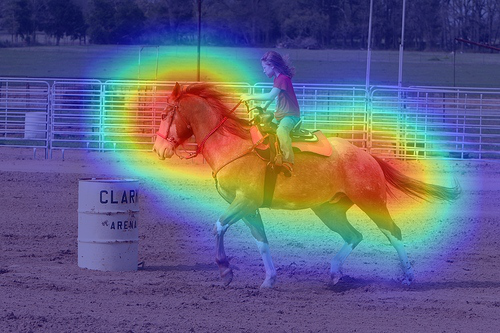}&\includegraphics[width=2cm]{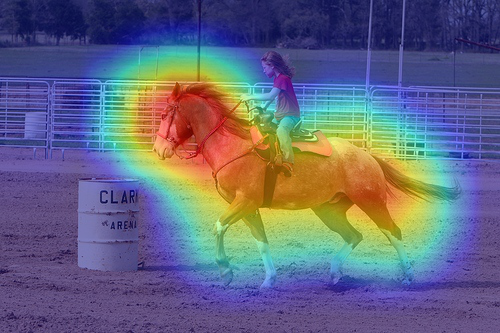} \\
\includegraphics[width=2cm]{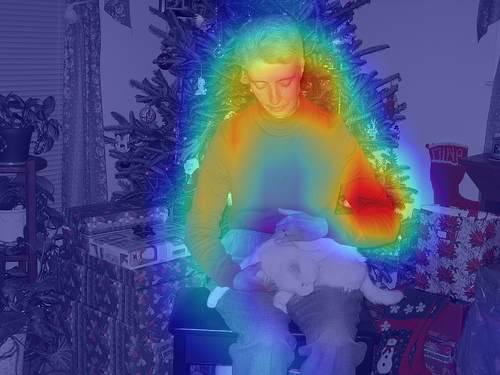}&\includegraphics[width=2cm]{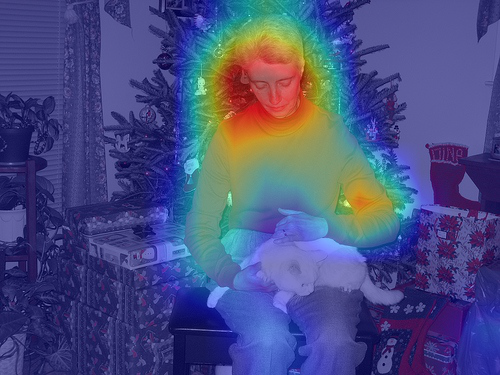}&\includegraphics[width=2cm]{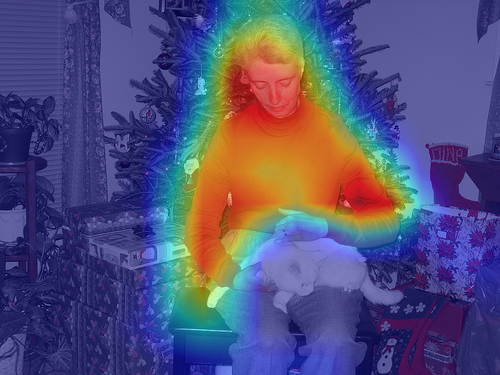}&
\includegraphics[width=2cm]{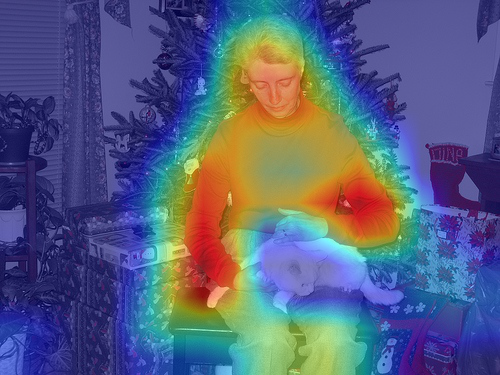} \\
$\alpha=0$&$\alpha = 0.01$ & $\alpha = 0.05$ & $\alpha = 0.1$ 
\end{tabular}
\end{adjustbox}
\caption{Attention maps obtained using Grad-CAMs from the end-to-end adversarial erasing method using different values for the adversarial loss term $\alpha$. As the $\alpha$ value increases, the attention spreads to less discriminative object regions.}
\label{fig:alphas}
\end{figure}
\endgroup

Our first experiment is an ablation study to verify our hypothesis that the adversarial network forces the localizer network to spread its attention to less discriminative object regions.
Recall from Equation~\ref{eq:totalloss} that $\alpha$ controls the strength of the adversarial loss term.
In Table~\ref{tab:alpha-ablation} we vary the $\alpha$ parameter and report the mIoU, precision and recall of the segmentation masks. 
We make the following observations: first, we find that a higher $\alpha$ indeed increases the recall, i.e. it forces the localizer to spread its attention to less discriminative object regions. 
Second, we observe that this increase in recall also increases performance in terms of mIoU.
The highest mIoU is obtained at  $\alpha = 0.05$, which strikes the right balance between precision and recall. 
A higher $\alpha$ value further increases the recall but the degradation in precision is stronger, resulting in a lower mIoU score.
Example attention maps generated for different $\alpha$ values are shown in Figure~\ref{fig:alphas}. 
Consistent with the previous observation, we see that a higher $\alpha$ value forces the attention map to spread to less discriminative object regions.
However, when the value is too high some pixels belonging to other classes and background regions receive high responses and therefore cause a drop in the precision.

A similar effect can be achieved by tuning the threshold ($\rho$) without needing an adversarial model. 
A lower threshold value increases the recall and decreases the precision, and vice versa.
However, decreasing the threshold value to obtain higher recall is unsatisfactory, as the localizer is not trained to find less discriminative regions belonging to the same object.
As a result, the localizer only focuses on the most discriminative object regions, failing to capture the entire object.

\subsection{Comparison to PSA}
\begin{table*}
\centering
\begin{tabular}{lccccc}
\toprule
      &      & CAM       &             & AffinityNet & DeepLabV3+    \\ \cmidrule(lr){2-4}\cmidrule(lr){5-5}\cmidrule(lr){6-6}
Model & mIoU & Precision & Recall      & mIoU         & mIoU \\ \midrule
PSA   & 46.8     &  60.3$^\dagger$        &   66.7$^\dagger$         &   58.7      &  60.7$^\dagger$    \\
PSA w/ EADER    & \textbf{48.6}     &   \textbf{61.3}        &   \textbf{68.7}          &  \textbf{60.1}            &  \textbf{62.8}    \\ \bottomrule
\end{tabular}
\caption{Comparison to our baseline, Pixel-level Semantic Affinity (PSA), on the Pascal VOC 2012 validation set. To enable a fair comparison we reproduce the PSA numbers and train the proxy labels from AffinityNet on DeepLabV3+. The numbers with a $\dagger$ denote our reproduced results.}
\label{table:baseline}
\end{table*}

\begingroup
\setlength{\tabcolsep}{1pt} 
\renewcommand{\arraystretch}{0.5} 
\begin{figure}[t!]
\centering
\begin{adjustbox}{max width=\hsize}
\begin{tabular}{c c c c}
\includegraphics[width=2.0cm]{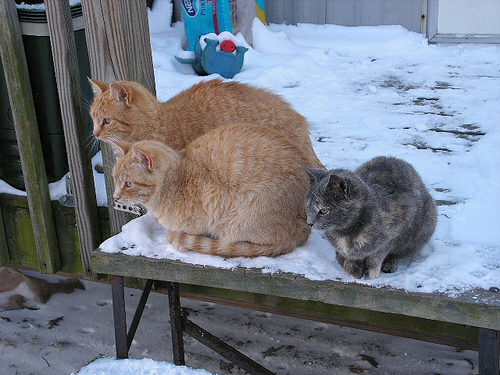}&
\includegraphics[width=2.0cm]{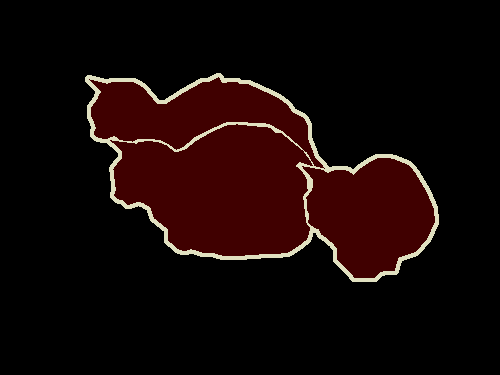}&
\includegraphics[width=2.0cm]{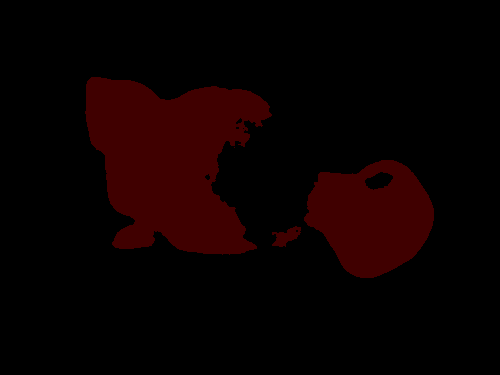}&
\includegraphics[width=2.0cm]{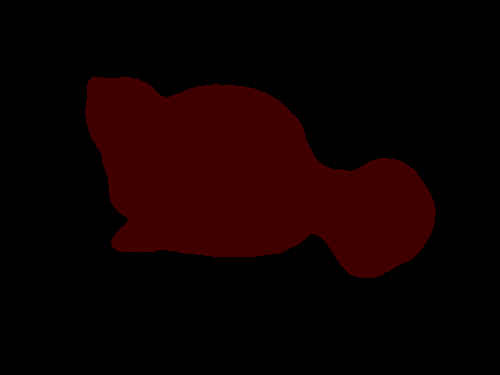}\\
\includegraphics[width=2.0cm]{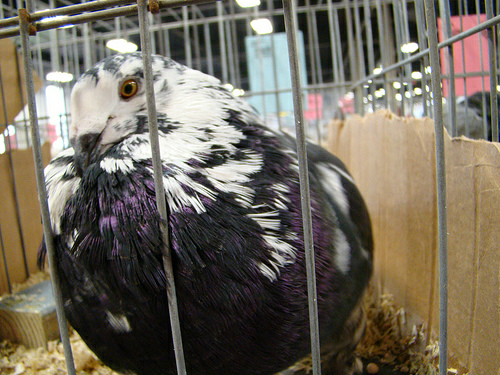}&
\includegraphics[width=2.0cm]{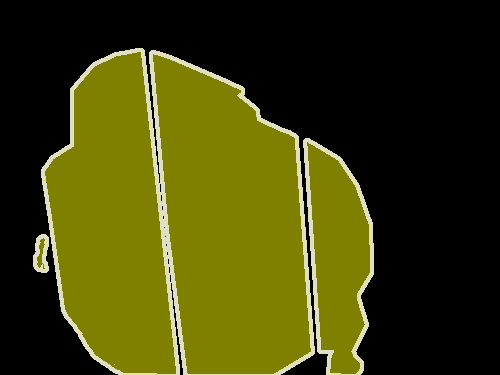}&
\includegraphics[width=2.0cm]{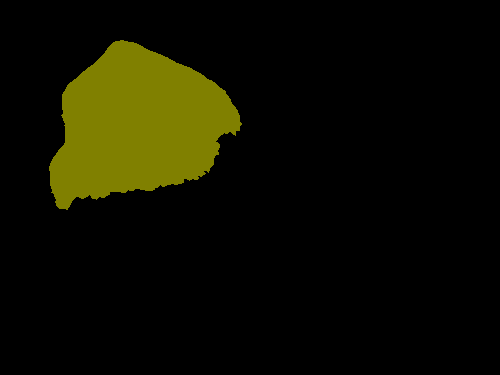}&
\includegraphics[width=2.0cm]{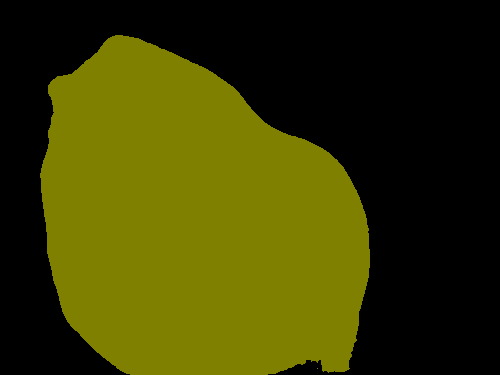}\\
\includegraphics[width=2.0cm]{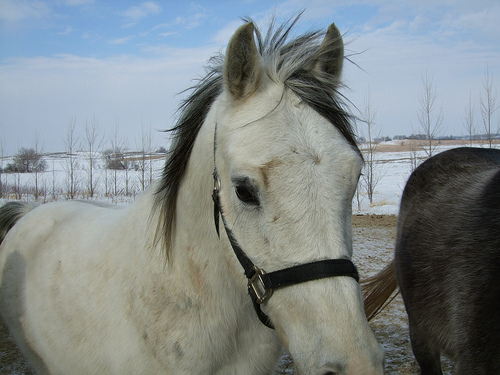}&
\includegraphics[width=2.0cm]{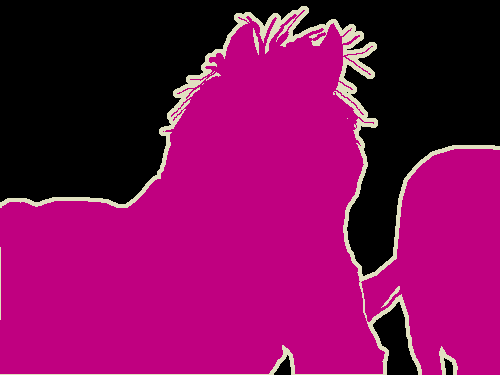}&
\includegraphics[width=2.0cm]{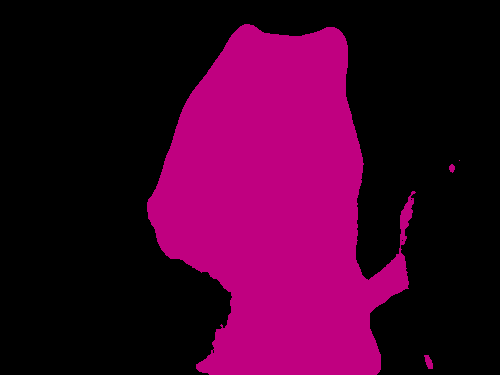}&
\includegraphics[width=2.0cm]{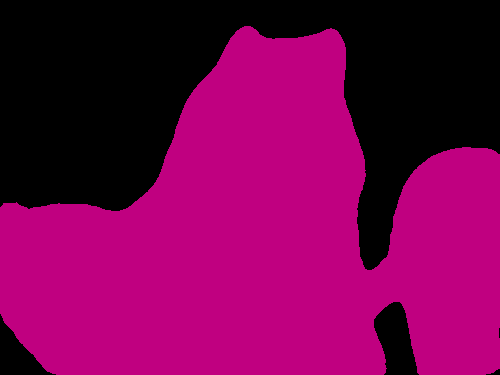}\\
\includegraphics[width=2.0cm]{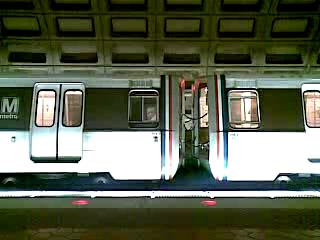}&
\includegraphics[width=2.0cm]{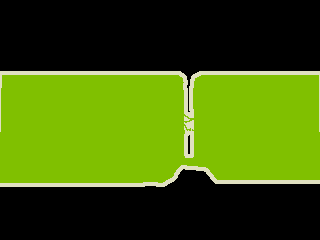}&
\includegraphics[width=2.0cm]{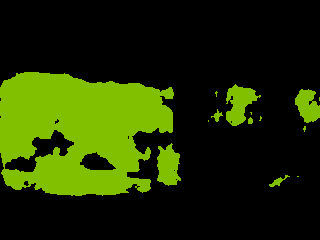}&
\includegraphics[width=2.0cm]{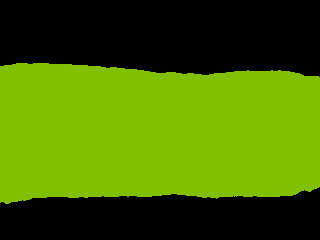}\\
\includegraphics[width=2.0cm]{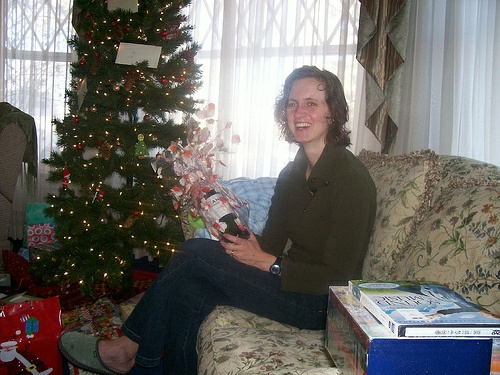}&
\includegraphics[width=2.0cm]{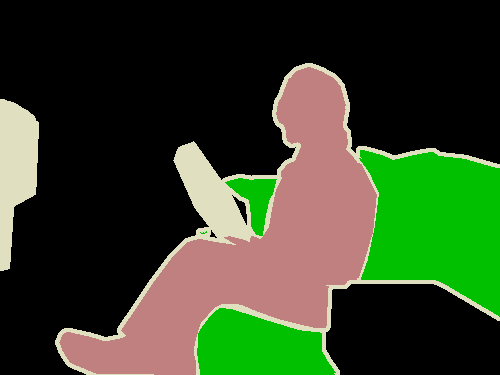}&
\includegraphics[width=2.0cm]{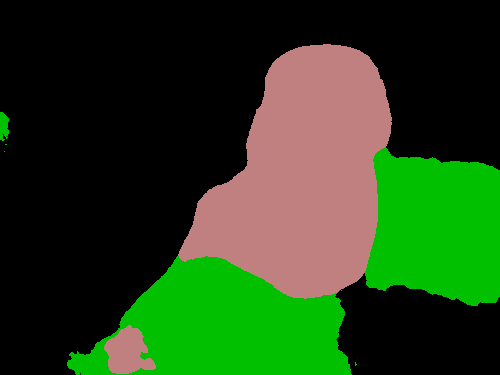}&
\includegraphics[width=2.0cm]{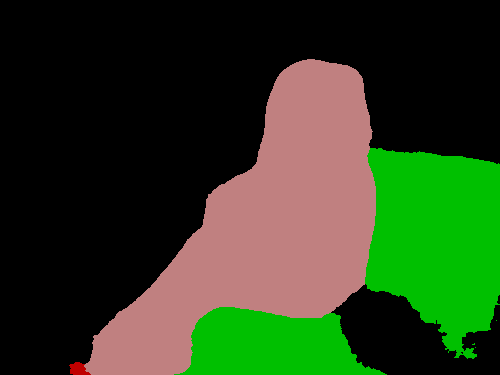}\\
\includegraphics[width=2.0cm]{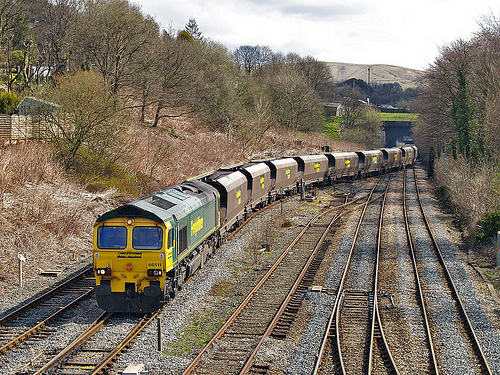}&
\includegraphics[width=2.0cm]{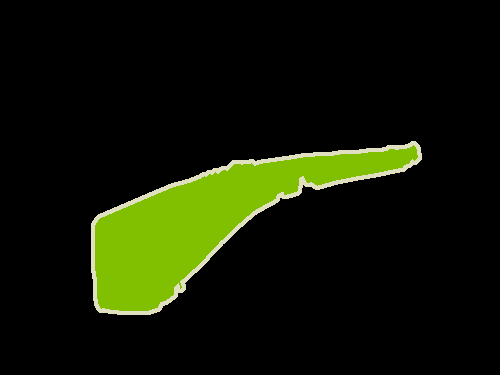}&
\includegraphics[width=2.0cm]{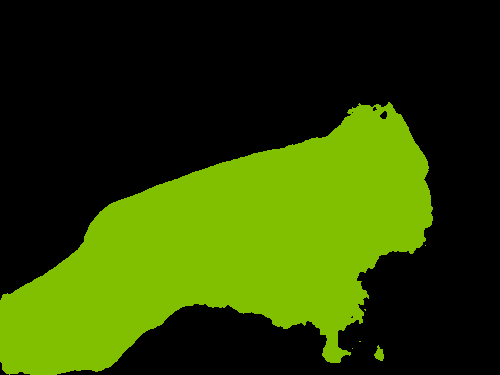}&
\includegraphics[width=2.0cm]{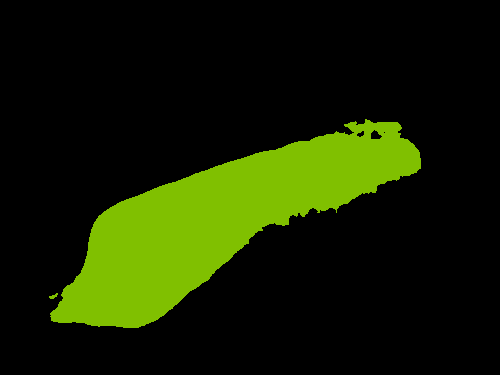}\\
Input & Ground Truth & PSA & PSA w/ EADER
\end{tabular}
\end{adjustbox}
\caption{Qualitative results on the Pascal VOC 2012 validation set. The white edges in the ground truth mask denote pixels that are ignored during evaluation. End-to-end adversarial erasing strategy increases the recall without sacrificing the precision.}
\label{fig:final}
\end{figure}
\endgroup

\begin{table*}[!t]
\centering
\begin{tabular*}{\textwidth}{l @{\extracolsep{\fill}} cccccccccccc}
\toprule
Method                              & bkg  & aero & bike & bird & boat & bottle & bus  & car  & cat  & chair & cow  \\ \midrule
PSA   & 86.7  & 53.2 & 29.1 & 76.7 & 44.2 & 67.7   & \textbf{85.2} & 72.4 & 71.7 & \textbf{26.7}  & 76.5  \\
PSA w/ EADER   & \textbf{88.2} & \textbf{54.9} & \textbf{31.3} & \textbf{84.1} & \textbf{58.2} & \textbf{70.9} & 83.0 & \textbf{76.2}      & \textbf{82.1}      & 24.4 & \textbf{80.6}
\end{tabular*}

\begin{tabular*}{\textwidth}{l @{\extracolsep{\fill}} cccccccccccc}
\toprule
Method                               & table & dog  & horse & mbike & person & plant & sheep & sofa & train & tv   & \colorbox{lightgray}{mean} \\ \midrule
PSA  &  \textbf{40.9}  & 72.2 & 68.2  & 70.2  & 66.4   & \textbf{37.8}  & \textbf{80.9}  & \textbf{38.5} & 62.8  & 45.4 & 60.7 \\
PSA w/ EADER    & 35.8      & \textbf{80.7} & \textbf{76.4} & \textbf{73.7}  & \textbf{70.8}     &  15.4      & 77.2  & 34.6      & \textbf{66.4}  & \textbf{52.6} & \textbf{62.8}\\
\bottomrule
\end{tabular*}
\caption{Per-class comparison with Pixel-level Semantic Affinity (PSA) on Pascal VOC 2012 validation set with only image-level supervision.}
\label{tab:class-comp-val}
\end{table*}

We now compare the original PSA results to the results where we have integrated end-to-end adversarial erasing.
Table~\ref{table:baseline} shows the improvements in terms of mIoU. 
Additionally, we report precision and recall after the CAM generation stage.
End-to-end adversarial erasing improves performance in this stage for all metrics.
In other words, the combination of the adversarial and regularization loss terms forces the attention map to spread to less discriminative object regions without spreading to background areas.
Besides, the mIoU scores in this stage are higher than those reported in Table~\ref{tab:alpha-ablation}, which is caused by the extensive test-time augmentations used by PSA.
In the next stage, training AffinityNet with the improved outputs of the first stage again results in better mIoU scores.
Finally, we report results when training a fully supervised semantic segmentation model on the proxy labels generated by AffinityNet. 
We report the results of training DeepLabV3+ on the proxy labels generated from PSA with and without end-to-end adversarial erasing.
Again, with end-to-end adversarial erasing the mIoU improves, showing the integrability of end-to-end adversarial erasing into existing WSSS methods. 
In Table~\ref{tab:class-comp-val} we make a per-class comparison of mIoU scores on the validation set.
Recall that the discriminative localization problem is especially prevalent in non-rigid object classes.
We find that end-to-end adversarial erasing significantly improves the results in many non-rigid object classes such as \textit{bird}, \textit{cat}, \textit{cow} and \textit{horse}.
Typically in these object classes the most discriminative object region is the head or the feet, which causes the attention map to cover only a small portion of these object classes.
With end-to-end adversarial erasing, the localizer is forced to capture the entire object region, as the fur or skin of these object classes are less discriminative, but still recognizable.
For outdoor object classes the results are often similar to PSA, while for indoor object classes the performance is often degraded.
Overall, end-to-end adversarial erasing increases the performance.

In Figure~\ref{fig:final} we show some qualitative results demonstrating the increase in precision, recall and mIoU. 
In the first four rows we find that end-to-end adversarial erasing better segments objects by capturing less discriminative object regions, especially for non-rigid object classes.
The increased specificity, as for instance observed in the last samples, can be attributed to the regularization term that forces the attention to spread only to areas where the localizer is confident that it is an object region.

\begin{table*}[!ht]
\centering
\begin{adjustbox}{max width=\textwidth}
\begin{tabular}{@{}lllccc@{}}
\toprule
Method   & Feature Extractor & Fully Supervised Model {(\scriptsize Backbone)}  & Supervision & Validation & Test \\ \midrule
FCN~\cite{long2015fully} & - & \scriptsize{(VGG16)} & $\mathcal{F}$ & - & 62.2 \\
WideResNet-38~\cite{wu2019wider} & - & \scriptsize{(WideResNet-38)} & $\mathcal{F}$ & 80.8 & 82.5 \\
DeepLabV3+~\cite{chen2018encoder} & - & \scriptsize{(Xception-65)} & $\mathcal{F}$          &  84.6          & 87.8      \\ \midrule
AE-PSL\cite{wei2017object} & VGG-16 & DeepLab {(\scriptsize VGG-16)} & $\mathcal{I} + \mathcal{S}$ & 55.0       & 55.7 \\
GAIN~\cite{li2018tell} & VGG-16 & DeepLab {(\scriptsize VGG-16)} & $\mathcal{I} + \mathcal{S}$           & 55.3       & 56.8 \\
SeeNet~\cite{hou2018self} & VGG-16 & DeepLab {(\scriptsize VGG-16)} & $\mathcal{I} + \mathcal{S}$ & 61.1 & 60.7 \\ 
FickleNet~\cite{lee2019ficklenet} & VGG-16 & DeepLab {(\scriptsize VGG-16)} & $\mathcal{I} + \mathcal{S}$ & 61.2 & 61.9 \\
Fan \emph{et al. }\cite{fan2018associating} & ResNet-50 & DeepLab {(\scriptsize VGG-16)} & $\mathcal{I} + \mathcal{S}$ & 61.3 & 62.1 \\
SeeNet~\cite{hou2018self} & VGG-16 & DeepLab {(\scriptsize ResNet-101)} &  $\mathcal{I} + \mathcal{S}$ & 63.1 & 62.8 \\ 
OAA+~\cite{jiang2019integral} & VGG-16 & DeepLab {(\scriptsize VGG-16)} & $\mathcal{I} + \mathcal{S}$ & 63.1 & 62.8 \\
Fan \emph{et al.}~\cite{fan2018associating} & ResNet-50 & DeepLab {(\scriptsize ResNet-101)} & $\mathcal{I} + \mathcal{S}$ & 63.6 & 64.5 \\
FickleNet~\cite{lee2019ficklenet} & VGG-16 & DeepLab {(\scriptsize ResNet-101)} & $\mathcal{I} + \mathcal{S}$ & 64.9 & 65.3 \\
OAA+~\cite{jiang2019integral} & VGG-16 & DeepLab {(\scriptsize ResNet-101)} & $\mathcal{I} + \mathcal{S}$ & 65.6 & 66.4 \\\midrule
EM-Adapt~\cite{papandreou2015weakly} & VGG-16 & - & $\mathcal{I}$   & 38.2 & 39.6 \\
SEC~\cite{kolesnikov2016seed} & VGG-16 & DeepLab {(\scriptsize VGG-16)} & $\mathcal{I}$  & 50.7 & 51.7 \\
MMEF~\cite{ge2018multi} & VGG-16 & FCN {(\scriptsize VGG-16)} & $\mathcal{I}$   & - & 55.6 \\
PSA~\cite{ahn2018learning}  (baseline) & WideResNet-38 & DeepLab {(\scriptsize VGG-16)}  & $\mathcal{I}$         & 58.4       & 60.5 \\
RRM~\cite{zhang2019reliability} & WideResNet-38 & DeepLab {(\scriptsize VGG-16)} & $\mathcal{I}$  & 60.7 & 61.0\\
PSA~\cite{ahn2018learning}  (baseline) & WideResNet-38 & WideResNet-38  & $\mathcal{I}$         & 61.7       & 63.7 \\
Araslanov and Roth~\cite{araslanov2020single} & WideResNet-38 & -  & $\mathcal{I}$ & 62.7 & 64.3 \\
IRNet~\cite{ahn2019weakly} & ResNet-50 & DeepLab  {(\scriptsize ResNet-50)}& $\mathcal{I}$ & 63.5 & 64.8 \\
SSDD~\cite{shimoda2019self} & WideResNet-38 & WideResNet-38 & $\mathcal{I}$  & 64.9 & 65.5 \\
RRM~\cite{zhang2019reliability} & WideResNet-38 & DeepLab {(\scriptsize ResNet-101)} & $\mathcal{I}$  & \textbf{66.3} & \textbf{66.5}\\ \midrule
PSA w/ EADER (Ours) & WideResNet-38 & DeepLab {(\scriptsize ResNet-101)} & $\mathcal{I}$   & 62.5 & 63.0  \\
PSA w/ EADER (Ours)   & WideResNet-38 & DeepLab {(\scriptsize Xception-65)} & $\mathcal{I}$     & 62.8       &  63.8     \\ \bottomrule
\end{tabular}
\end{adjustbox}
\caption{Comparison of WSSS methods on the Pascal VOC 2012 dataset. For the supervision, $\mathcal{I}$ denotes image-level labels, $\mathcal{S}$ denotes saliency masks and $\mathcal{F}$ denotes pixel-level labels, which is the upper bound for fully supervised semantic segmentation. For the feature extractor, the mentioned architecture is the one that is used to generate the initial object locations (e.g. by utilizing CAMs).
}
\label{tab:sota}
\end{table*}

\subsection{Comparison to Adversarial Erasing Methods}
\begin{table}[t]
\centering
\begin{adjustbox}{max width=\textwidth}
\begin{tabular}{@{}lccc@{}}
\toprule
Method   &  Supervision & Validation & Test \\ \midrule
AE-PSL\cite{wei2017object} & $\mathcal{I} + \mathcal{S}$ & 55.0       & 55.7 \\
GAIN~\cite{li2018tell}  & $\mathcal{I} + \mathcal{S}$           & 55.3       & 56.8 \\
SeeNet~\cite{hou2018self}  &  $\mathcal{I} + \mathcal{S}$ & \textbf{63.1} & 62.8 \\ \midrule
ACoL~\cite{zhang2018adversarial} & $\mathcal{I}$ & 56.1$^\dagger$ & - \\ \midrule
PSA w/ EADER (Ours) & $\mathcal{I}$ & 62.8       &  \textbf{63.8}    \\ \bottomrule
\end{tabular}
\end{adjustbox}
\caption{Comparison with the previous adversarial erasing methods for WSSS on the Pascal VOC 2012 dataset. For supervision, $\mathcal{I}$ denotes image-level labels and $\mathcal{S}$ denotes saliency masks. The result with $\dagger$ was obtained from~\cite{hou2018self}.
}
\label{tab:adv-erasing}
\end{table}
In Table~\ref{tab:adv-erasing} we compare our results to previous WSSS methods that follow an adversarial erasing strategy.
Note that in each method the adversarial erasing is a component in a multi-stage setup.
We outperform all existing adversarial erasing methods, even when most of them use stronger supervision signals in the form of saliency masks.
When comparing to ACoL~\cite{zhang2018adversarial}, the only other adversarial erasing methodology without saliency masks, we significantly outperform their method on the validation set.

\subsection{Comparison to the State-of-the-Art}
In Table~\ref{tab:sota} we compare to the previous WSSS methods, where we report the feature extractor that is used to generate object locations and the fully supervised model that is trained on proxy labels.
The methods denoted with supervision signal $\mathcal{F}$ denote the upper bound of the segmentation performance.
Note that our method outperforms a fully supervised FCN~\cite{long2015fully} network and we achieve $\approx 73\%$ of the upper bound, set by a fully supervised DeepLabV3+\cite{chen2014semantic} model.
Further, we show that we outperform many existing WSSS methods, but are outperformed by some others.

\section{Conclusion}
In this paper we presented a novel end-to-end adversarial erasing method to resolve the discriminative localization problem, an inherent issue in weakly-supervised semantic segmentation methods. 
This approach is easily integrable to existing methods, not requiring iterative classifiers, post-processing, weight sharing or saliency masks, unlike many previous adversarial erasing methods.
We further show that end-to-end adversarial erasing improves performance on the Pascal VOC 2012 dataset, especially on most non-rigid object classes, which suffer the most from the discriminative localization problem.

{\small
\bibliographystyle{ieee_fullname}
\bibliography{bib}
}

\end{document}